\documentclass[11pt]{article}
\usepackage{eacl2017}
\usepackage{times}
\usepackage{graphicx}
\usepackage{url}
\usepackage{latexsym}
\usepackage{multirow}
\usepackage{caption}
\usepackage{subcaption}
\usepackage{tabularx}
\usepackage[linesnumbered]{algorithm2e}

\eaclfinalcopy 

\title{Question Generation from a Knowledge Base with Web Exploration}

\author{Linfeng Song$^1$ \and Lin Zhao$^2$ \\
  $^1$Computer Science Department, University of Rochester, Rochester, NY, 14623 \\
  $^2$Bosch Research and Technology Center, Palo Alto, CA, 94304  \\
  }

\date{}

\begin{document}
\maketitle
\begin{abstract}
Question generation from a knowledge base (KB) is the task of generating questions related to the domain of the input KB.
We propose a system for generating fluent and natural questions from a KB, which significantly reduces the human effort by leveraging massive web resources.
In more detail, a seed question set is first generated by applying a small number of hand-crafted templates on the input KB, 
then more questions are retrieved by iteratively forming already obtained questions as search queries into a standard search engine,
before finally questions are selected by estimating their fluency and domain relevance.
Evaluated by human graders on 500 random-selected triples from Freebase, questions generated by our system are judged to be more fluent than those of \newcite{serban-EtAl:2016:P16-1} by human graders.
\end{abstract}

\section{Introduction}

Question generation is important as questions are useful for student assessment or coaching purposes in educational or professional contexts, and a large-scale corpus of question and answer pairs is also critical to many NLP tasks including question answering, dialogue interaction and intelligent tutoring systems.
There has been much literature so far
\cite{chen2009aist,ali2010automation,heilman2010good,curto2012question,lindberg2013generating,mazidi2014linguistic,labutov2015deep} studying question generation from text.
Recently people are becoming interested in question generation from KB, since large-scale KBs, such as Freebase \cite{bollacker2008freebase} and DBPedia \cite{auer2007dbpedia}, are freely available, and entities and their relations are already present in KBs but not for texts.

Question generation from KB is challenging as function words and morphological forms for entities are abstracted away when a KB is created.
To tackle this challenge, previous work \cite{seyler2015generating,serban-EtAl:2016:P16-1} relies on massive human-labeled data.
Treating question generation as a machine translation problem, \newcite{serban-EtAl:2016:P16-1} train a neural machine translation (NMT) system with 10,000 $\langle$triple\footnote{A triple is a $\langle$subject,predicate,object$\rangle$ in KB, such as $\langle$jigsaw, performsActivity, CurveCut$\rangle$}, question$\rangle$ pairs.
At test time, input triples are ``translated'' into questions with the NMT system.
On the other hand, the question part of the 10,000 pairs are human generated, which requires a large amount of human effort.
In addition, the grammaticality and naturalness of generated questions can not be guaranteed (as seen in Table \ref{tab:30m}).

We propose a system for generating questions from KB that significantly reduces the human effort by leveraging the massive web resources.
Given a KB, a small set of question templates are first hand-crafted based on the predicates in the KB.
These templates consist of a transcription of the predicate in the KB (e.g. performsActivity$\Rightarrow$how to) and placeholders for the subject (\#X\#) and the object (\#Y\#).
A seed question set is then generated by applying the templates on the KB.
The seed question set is further expanded through a search engine (e.g., Google, Bing), by iteratively forming each generated question as a search query to retrieve more related question candidates. 
Finally a selection step is applied by estimating the fluency and domain relevance of each question candidate.

The only human labor in this work is the question template construction.
Our system does not require a large number of templates because: (1) the iterative question expansion can produce a large number of questions even with a relatively small number of seed questions, as we see in the experiments, (2) multiple entities in the KB share the same predicates.
Another advantage is that our system can easily generate updated questions as web is self-updating consistently.
In our experiment, we compare with \newcite{serban-EtAl:2016:P16-1} on 500 random selected triples from Freebase \cite{bollacker2008freebase}.
Evaluated by 3 human graders, questions generated by our system are significantly better then \newcite{serban-EtAl:2016:P16-1} on grammaticality and naturalness.

\section{Knowledge Base}

A knowledge base (KB) can be viewed as a directed graph, in which nodes are entities (such as ``jigsaw'' and ``CurveCut'') and edges are relations of entities (such as ``performsActivity'').
A KB can also be viewed as a list of triples in the format of $\langle$subject, predicate, object$\rangle$, 
where subjects and objects are entities, and predicates are relations.

\section{System}

\begin{figure}
\centering
\includegraphics[scale=.6]{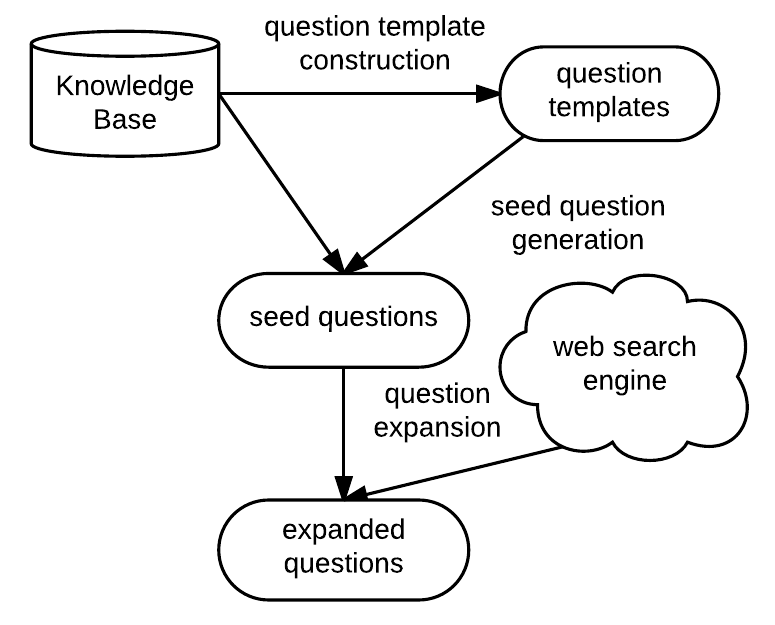}
\caption{Overview of our framework.}
\label{fig:overview}
\end{figure}

\begin{figure}
\centering
\includegraphics[width=\columnwidth]{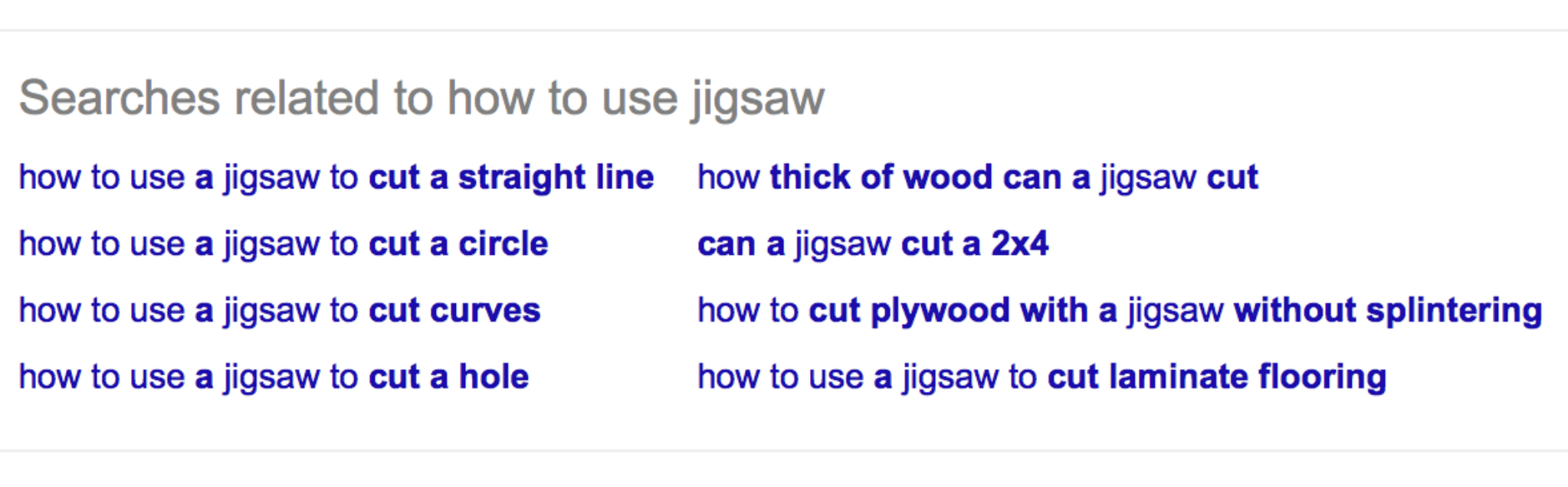}
\caption{Related search results for the question ``how to use jigsaw''.}
\label{fig:google}
\end{figure}

Shown in Figure \ref{fig:overview}, our system contains the sub-modules of question template construction, seed question generation, question expansion and selection.
Given an input KB, a small set of question templates is first constructed such that each template is associated with a predicate,
then a seed question set is generated by applying the template set on the input KB, before finally more questions are generated from related questions that are iteratively retrieved from a search engine with already-obtained questions as search queries (section \ref{sec:exp}).
Taking our in-house KB of power tool domain as an example, template ``how to use \#X\#'' is first constructed for predicate ``performsActivity''.
In addition, seed question ``how to use jigsaw'' is generated by applying the template on triple ``$\langle$jigsaw, performsActivity, CurveCut$\rangle$'', before finally questions (Figure \ref{fig:google}) are retrieved from Google with the seed question.

\subsection{Question expansion and selection}
\label{sec:exp}

\begin{algorithm}[t]
 \KwData{seed question set $S$}
 \KwResult{candidate questions $E$}
 $E \leftarrow S$\;
 $Q \leftarrow S$\;
 $I \leftarrow 0$\;
 \While{\textsc{len}$(Q) > 0$ \textbf{and} $I < I_{max}$}{
    $I = I + 1$\;
    $q_{cur}$ $\leftarrow$ $Q$.\textsc{Pop}()\;
    \For{$q_{next}$ \textbf{in} \textsc{WebExp}$(q_{cur})$}{
       \If{\textbf{not} $E$.\textsc{contains}$(q_{next})$}{
           $E$.\textsc{Append}($q_{next}$) \;
           $Q$.\textsc{Push}($q_{next}$)\;
       }
    }
 }
 \caption{Question expansion method}
 \label{algo:exp}
\end{algorithm}

Shown in Algorithm \ref{algo:exp}, the expanded question set $E$ is initialized as the seed question set (Line 1).
In each iteration, an already-obtained question is expanded from web and the retrieved questions are  added to $E$ if $E$ does not contain them (Lines 6-10).
As there may be a large number of questions generated in the loop, we limit the maximum number of iterations with $I_{max}$ (Line 4).

\begin{table*} \small
\centering
\begin{tabularx}{\textwidth}{X|X} 
\hline
Ours & \newcite{serban-EtAl:2016:P16-1} \\
\hline
\hline
what is the cultural heritage of churchill national park & where in australia is churchill national park \\
\hline
what percentage of argentina's population live in urban areas & what 's one of the mountain where can you found in argentina in netflix \\
\hline
which country is the largest financial center of latin america & what is an organization that was born in latin america \\
\hline
which country has the largest freshwater lake in central america & what are the major town three gringos in venezuela and central america book \\
\hline
how does leukemia affect the body in children & who was someone who was involved in the leukemia \\
\hline
how does the nervous system maintain homeostasis & what is the drug category of central nervous system stimulation \\
\hline
why were colonial minutemen so prepared for the arrival of the british in concord & what county is concord in \\
\hline
which is the only country to have a bible on their national flag & whats the title of a book of the subject of the bible \\
\hline
why is new york called the city that never sleeps & who was born in new york \\
\hline
what three important documents were written in pennsylvania & what is located in pennsylvania \\
\hline
\end{tabularx}
\caption{Comparing generated questions}
\label{tab:30m}
\end{table*}

\begin{table} \small
\centering
\begin{tabular}{|c|c|c|}
\hline
System & grammatical & naturalness \\
\hline
\hline
\newcite{serban-EtAl:2016:P16-1} & 3.36 & 3.14 \\
Ours & \textbf{3.53} & \textbf{3.31} \\
\hline
\end{tabular}
\caption{Human ratings of generated questions}
\label{tab:30m_quan}
\end{table}

The questions collected from the web search engine may not be fluent or domain relevant; especially the domain relevance drops significantly as the iteration goes on.
Here we adopt a skip-gram model \cite{mikolov2013distributed} and a language model for evaluating the domain relevance and fluency of the expanded questions, respectively.
For domain relevance, we take the seed question set as the in-domain data $D_{in}$,
the domain relevance of expanded question $q$ is defined as:
\begin{equation}
\textsc{Rel}(q) = \cos(v(q),v(D_{in}))
\label{eq:rel}
\end{equation}
where $v(\cdot)$ is the document embedding defined as the averaged word embedding within the document.
For fluency, we define the averaged language model score as:
\begin{equation}
\textsc{AvgLM}(q) = \frac{\textsc{Lm}(q)}{\textsc{Len}(q)}
\label{eq:lm}
\end{equation}
where $\textsc{Lm}(\cdot)$ is the general-domain language model score (log probability), and $\textsc{Len}(\cdot)$ is the word count.
We apply thresholds $t_{rel}$ and $t_{flu}$ for domain relevance and fluency respectively, and 
filter out questions whose scores are below these thresholds.

\section{Experiments}

We perform three experiments to evaluate our system qualitatively and quantitatively. 
In the first experiment, we compare our end-to-end system with the previous state-of-the-art method \cite{serban-EtAl:2016:P16-1} on Freebase \cite{bollacker2008freebase}, a domain-general KB.
In the second experiment, we validate our domain relevance evaluation method on a standard dataset about short document classification. 
In the final experiment, we run our end-to-end system on a highly specialized in-house KB and present sample results, showing that our system is capable of generating questions from domain specific KBs.

\subsection{Evaluation on Freebase}

We first compare our system with \newcite{serban-EtAl:2016:P16-1} on 500 randomly selected triples from Freebase \cite{bollacker2008freebase}\footnote{We obtain their results from http://agarciaduran.org}.
For the 500 triples, we hand-crafted 106 templates, as these triples share only 53 distinct predicates (we made 2 templates for each predicate on average).
991 seed questions are generated by applying the templates on the triples, and 1529 more questions are retrieved from Google.
To evaluate the fluency of the candidate questions, we train a 4-gram language model (LM) on gigaword (LDC2011T07) with Kneser Ney smoothing.
Using the averaged language model score as index, the top 500 questions are selected to compare with the results from \newcite{serban-EtAl:2016:P16-1}.
We ask three native English speakers to evaluate the fluency and the naturalness\footnote{whether people will ask in reality} of both results based on a 4-point scheme where 4 is the best.

We show the averaged human rate in Table \ref{tab:30m_quan}, where we can see that our questions are more grammatical and natural than \newcite{serban-EtAl:2016:P16-1}. The naturalness score is less than the grammatical score for both methods. It is because naturalness is a more strict metric since a natural question should also be grammatical.

Shown in Table \ref{tab:30m}, we compare our questions with \newcite{serban-EtAl:2016:P16-1} where questions in the same line describe the same entity.
We can see that our questions are grammatical and natural as these questions are what people usually ask on the web.
On the other hand, questions from \newcite{serban-EtAl:2016:P16-1} are either ungrammatical (such as ``who was someone who was involved in the leukemia ?'' and ``whats the title of a book of the subject of the bible ?''), 
unnatural (``what 's one of the mountain where can you found in argentina in netflix ?'') or confusing (``who was someone who was involved in the leukemia ?'').

\begin{table}[t]
\centering
\begin{tabular}{|l|l|}
\hline
Method & Precision \\
\hline
\hline
\newcite{phan2008learning} & 82.18 \\
\newcite{chen2011short} & 85.31 \\
\newcite{ma-EtAl:2015:VSM-NLP} & 85.48 \\
Ours & \textbf{85.65} \\
\hline
\end{tabular}
\caption{Precision on the web snippet dataset}
\label{tab:relevance}
\end{table}

\subsection{Domain Relevance}

We test our domain-relevance evaluating method on the web snippet dataset, which is a commonly-used for domain classification of short documents.
It contains 10,060 training and 2,280 test snippets (short documents) in 8 classes (domains), and each snippet has 18 words on average.
There have been plenty of prior results \cite{phan2008learning,chen2011short,ma-EtAl:2015:VSM-NLP} on the dataset.

Shown in Table \ref{tab:relevance}, we compare our domain-relevance evaluation method (section \ref{sec:exp}) with previous state-of-the-art methods: 
\newcite{phan2008learning} first derives latent topics with LDA \cite{blei2003latent} from Wikipedia, then uses the topics as appended features to expand the short text. 
\newcite{chen2011short} further expanded \newcite{phan2008learning} by using multi-granularity topics.
\newcite{ma-EtAl:2015:VSM-NLP} adopts a Bayesian model that the probability a document $D$ belongs to a topic $t$ equals to the prior of $t$ times the probability each word $w$ in $D$ comes from $t$. 
Our method first concatenates training documents of the same domain into one ``domain document'', then calculates each document embedding by averaging word embeddings within it, before finally assigns the label of the nearest (cosine similarity) ``domain document'' to each test document.

Simple as it is, our method outperforms all previous methods proving its effectiveness.
The reason can be that word embeddings captures the similarity between distinct words (such as ``finance'' and ``economy''), while it is hard for traditional methods.
On the order hand, LDA only learns probabilities of words belonging to topics.

\subsection{Evaluation on the Domain-specific KB}

The last experiment is on our in-house KB in the power tool domain.
It contains 67 distinct predicates, 293 distinct subjects and 279 distinct objects respectively.
For the 67 predicates, we hand-craft 163 templates.
Here we use the same language model as in our first experiment, and learn a skip-gram model \cite{mikolov2013distributed} on Wikipedia\footnote{https://dumps.wikimedia.org/} for evaluating domain relevance.

We generate 12,228 seed questions from which 20,000 more questions are expanded with Google.
Shown in Table \ref{tab:questions} are some expanded questions from which we can see that most of them are grammatical and relevant to the power tool domain.
In addition, most questions are informative and correspond to a specific answer, except the one ``do I need a hammer drill'' that lacks context information.
Finally, in addition to the simple factoid questions, our system generates many complex questions such as ``how to cut a groove in wood without a router''.

\begin{table}
\centering
\begin{tabular}{|l|}
\hline
how to change circular saw blade \\
how to measure lawn mower cutting height \\
how to sharpen drill bits on bench grinder \\
how does an oscillating multi tool work \\
how to cut a groove in wood without a router \\
what type of sander to use on deck \\
do i need a hammer drill \\
can i use acrylic paint on wood \\
how to use a sharpening stone with oil \\
\hline
\end{tabular}
\caption{Example question expanded}
\label{tab:questions}
\end{table}

\section{Conclusion}

We presented a system to generate natural language questions from a knowledge base. 
By leveraging rich web information, our system is able to generate domain-relevant questions in wide scope, while human effort is significantly reduced. 
Evaluated by human graders, questions generated by our system are significantly better than these from \newcite{serban-EtAl:2016:P16-1} on 500 random-selected triples from Freebase.
We also demonstrated generated questions from our in-house KB of power tool domain, which are fluent and domain-relevant in general.
Our current system only generates questions without answers, leaving automatic answer mining as our future work.



\bibliography{eacl2017}
\bibliographystyle{eacl2017}

\end{document}